\pdfoutput=1

\documentclass[11pt]{article}

\usepackage[preprint]{acl}

\usepackage{times}
\usepackage{latexsym}

\usepackage{colortbl}  
\usepackage{xcolor}    
\definecolor{babycyan}{rgb}{0.85, 0.97, 0.85}

\definecolor{aliceblue}{RGB}{255, 238, 241}

\usepackage{booktabs}

\usepackage[T1]{fontenc}

\usepackage[utf8]{inputenc}

\usepackage{microtype}

\usepackage{amsmath} 
\usepackage{multirow}
\usepackage{inconsolata}

\usepackage{graphicx}
\usepackage{amsfonts}
\usepackage{enumitem}
\usepackage{algorithm}
\usepackage{algpseudocode}
\usepackage{listings}
\usepackage{longtable}
\usepackage[listings,skins,breakable]{tcolorbox}
\title{EvolveSearch: An Iterative Self-Evolving Search Agent}

\newcommand{\Ours}{EvolveSearch}
\newcommand{\email}[1]{\href{mailto:#1}{\textcolor{black}{\texttt{#1}}}}

\author{%
Dingchu Zhang$^{*}$, Yida Zhao\thanks{These authors contributed equally.}, Jialong Wu, Baixuan Li, Wenbiao Yin, \\
\textbf{Liwen Zhang\hspace{0.5mm}~\footnotemark[2], 
 Yong Jiang\hspace{0.5mm}~\footnotemark[2], Yufeng Li\thanks{Correspondence.}, Kewei Tu, Pengjun Xie, Fei Huang} \\
  Tongyi Lab, Alibaba Group\\
     Correspondence to: \email{zhangdc@lamda.nju.edu.cn} \\
    \email{\{zlw439616,yongjiang.jy\}@alibaba-inc.com}
}

\begin{document}
\maketitle
\begin{abstract}


The rapid advancement of large language models (LLMs) has transformed the landscape of agentic information seeking capabilities through the integration of tools such as search engines and web browsers.
However, current mainstream approaches for enabling LLM web search proficiency face significant challenges: supervised fine-tuning struggles with data production in open-search domains, while RL converges quickly, limiting their data utilization efficiency. 
To address these issues, we propose \textbf{\Ours}, a novel iterative self-evolution framework that combines SFT and RL to enhance agentic web search capabilities without any external human-annotated reasoning data. 
Extensive experiments on seven multi-hop question-answering (MHQA) benchmarks demonstrate that \Ours\ consistently improves performance across iterations, ultimately achieving an average improvement of 4.7\% over the current state-of-the-art across seven benchmarks, opening the door to self-evolution agentic capabilities in open web search domains.

\end{abstract}

\section{Introduction}



Rapid advances in large language models (LLMs) have enabled agentic AI capabilities through tool integration (\textit{e.g.}, search, browsing, code execution), supporting autonomous interaction with external environments. 
Recent agentic systems like OpenAI \textit{Deep Research}~\cite{dr} achieve 51.9\% accuracy on BrowseComp, well above human performance (29.8\%)~\cite{wei2025browsecomp}, highlighting LLMs’ potential for deep information research.

\begin{figure}[t]
  \includegraphics[width=\columnwidth]{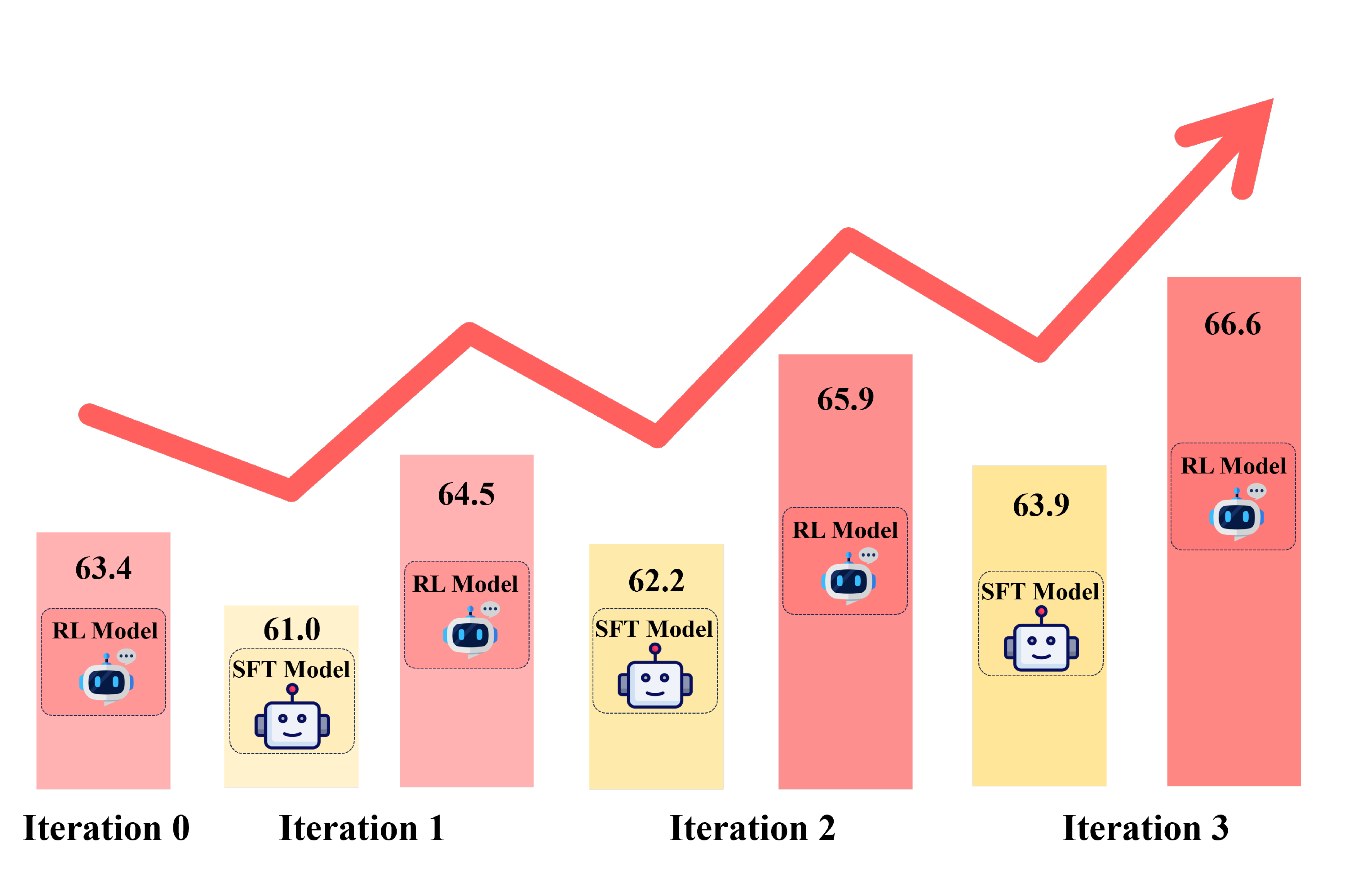}
  \caption{Iterative improvements in the average performance of SFT and RL models, reflecting progressive enhancement through self-evolution.}
  \label{Illustration}
\end{figure}

Existing agentic systems are primarily implemented via prompting-based, supervised fine-tuning (SFT)-based, and reinforcement learning (RL)-based approaches.
Prompting-based agents rely on predefined workflows~\cite{bulid,zhou2023agents}, resulting in rigid behaviors and limited generalization.
They often struggle with instruction following and reasoning, requiring substantial manual prompt engineering for reliable performance~\cite{pan2025multiagent}. 
Subsequent work distills agentic capabilities into smaller LLMs via supervised fine-tuning (SFT)~\cite{wang2025chain,wu2025webdancer}.
However, in open-ended search tasks, collecting SFT data necessitates complex environment interactions, making data construction challenging.
More importantly, they lack robustness in complex, real-world environments~\cite{zheng2025deepresearcher}.
RL-based approaches~\cite{shao2024deepseekmath} have recently gained attention for enabling models to acquire decision-making capabilities through online interactions with the environment and reward-driven updates.
This paradigm allows agents to adapt to task dynamics in an end-to-end manner.
However,  in practice, existing RL-based methods often converge within fewer than 100 steps, resulting in low data efficiency and limited performance gains~\cite{R1-searcher,jin2025search}.

To address the challenges of scarce SFT data and the limited performance of the existing RL approach, we propose \textbf{\Ours}, a novel iterative self-evolution framework that combines SFT and RL to enhance web search capabilities \textbf{without} any external human-annotated reasoning data. 
Specifically, \Ours\ proceeds in alternating phases:
(i) In the \underline{RL exploration} phase,  the model interacts with a web search environment, leveraging tool-use capabilities and receiving a hybrid reward signal. 
This enables the model to identify and learn from high-reward rollouts.
(ii) In the \underline{SFT optimization} phase, the best-performing rollouts from the RL phase are selected based on three criteria and used to optimize the model via SFT, yielding a stronger initialization (\textit{i.e.}, cold-start policy) for the next RL cycle.
By iteratively alternating between exploration and optimization, \Ours\ progressively bootstraps the performance of the RL model and the SFT model, learning robust and effective search behaviors from its own experience without human intervention, as illustrated in Figure~\ref{Illustration}.

To validate the effectiveness of \Ours, we conduct extensive experiments in realistic web search settings.
Evaluation on seven multi-hop question-answering (MHQA) benchmarks demonstrates \Ours\ consistently outperforms competitive baselines. 
Notably, it achieves a 4.7\% average accuracy gain over the previous state-of-the-art (SOTA), demonstrating the benefits of combining supervised fine-tuning and reinforcement learning in a self-evolution framework. 
These results highlight the strength of iterative learning from high-reward rollouts, enabling substantial performance improvements without reliance on human-annotated reasoning data.



Overall, our contributions are as follows:

\begin{itemize}[itemsep=0.6pt, parsep=0.1pt]



\item[$\bullet$] We propose \Ours, a novel framework that, to the best of our knowledge, is the first to iteratively combine RL with SFT to enhance LLMs' capabilities in the web search scenario.

\item[$\bullet$]  \Ours\ requires no human-annotated reasoning data; instead, it leverages high-quality rollouts from RL models to enable continuous self-improvement via self-generated supervision.

\item[$\bullet$] We conduct extensive empirical evaluations on multiple MHQA datasets, demonstrating the significant effectiveness and generality of \Ours\ over existing SOTA.

\end{itemize}

\section{Background}

In this work, a question-answering rollout expands through a ReAct-based~\cite{yao2022react} sequence of \texttt{thought}-\texttt{action}-\texttt{observation} iterations.
Within each iteration, the LLM agent generates: (i) A free-form \texttt{thought} ($\tau$) to extract information, adjust action plans, and track task progress, etc.
(ii) A structured \texttt{action} ($\alpha$) to interact with external environments.
(iii) This interaction yields an \texttt{observation} ($o$) which serves as feedback for the next iteration. 
Formally, we represent the agentic execution loop at a given time step $t$ as a triplet $(\tau_t, \alpha_t, o_t)$. 
In \Ours, the action $\alpha$ can be either \textit{search}, corresponding to the utilization of a search tool, or \textit{answer}, which involves formulating a response to the given question. The \texttt{observation} $o$ after a \textit{search} action typically includes a list of relevant results, such as the top-10 titles and snippets retrieved from the search tool. Consequently, the historical rollout leading up to time step $t$ denoted as, $\mathcal{H}_t$, can be represented as the sequence:
\begin{align}
\mathcal{H}_t=(\tau_0,\alpha_0,o_0,\tau_1,...,\tau_{t-1},\alpha_{t-1},o_{t-1}).
\end{align}
At time step $t$, the agent considers the historical rollout $\mathcal{H}_t$ to generate \texttt{thought} $\tau_t$ and subsequently select an \texttt{action} $\alpha_t$, following policy
$\pi(\tau_t,\alpha_t|\mathcal{H}_t)$. Then it gets a feedback \texttt{observation} $o$ if $\alpha_t$ is a \textit{search} action. Otherwise, the rollout comes to an end after the \textit{answer} action $\alpha_t$ is completed. 

\section{Method}

\begin{figure*}[t]
  \includegraphics[width=\linewidth]{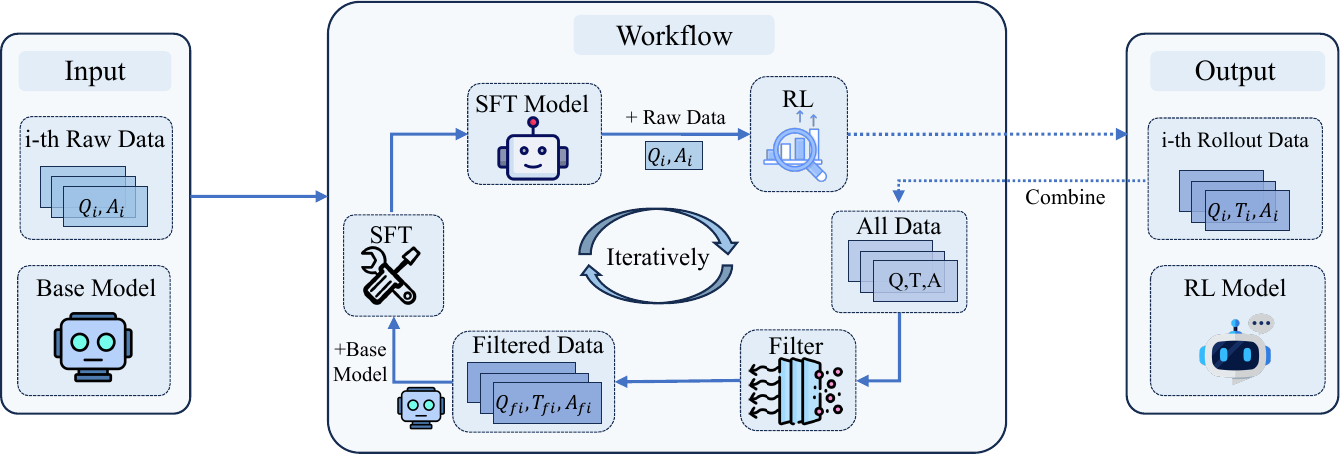}
  \caption{The overall framework of \textbf{\Ours}. 
  We iteratively input raw question-answering data and a fixed base model. 
  In the $i$-th iteration, the base model first performs SFT with a filter pool of training data, followed by RL with the $i$-th raw data.
  The rollout data during RL is filtered and  incorporated into the training data pool for SFT in the subsequent iteration.
  }
  \label{method}
\end{figure*}



\textbf{EvolveSearch} employs iterative Reinforcement Learning (RL) and Rejection Sampling Fine-tuning (RSFT) to train an LLM agent towards solving question-answering problems with multiturn thinking and tool use. We provide an overview of our framework, with the full workflow illustrated in Figure~\ref{method}. We begin by evenly dividing the dataset into $N$ parts. The process then involves iteratively executing the following two stages $N$ times:


\begin{itemize}
    \item \textbf{Stage 1}. 
    In the i-th iteration, we utilize the i-th portion of the raw data to perform RL training on the SFT model from the previous iteration. By implementing a hybrid reward mechanism, we obtain an RL model with better generalization capabilities and rollouts during training. These rollouts provide diverse and high-quality training data for the subsequent SFT training phase. 

    
    \item \textbf{Stage 2}. We merge the high-quality and diverse rollouts from the previous stage into the data pool, then refine the cold-start data by applying three filtering rules. This refined data is subsequently used for SFT of the fixed base model, thereby enabling a more robust cold-start model for RL training in the next iteration.
    
\end{itemize}

We will further explain these two stages in Section~\ref{rl} and Section~\ref{rsft}, respectively. The iterative algorithm workflow is presented in Appendix~\ref{Algorithm}.

\subsection{Hybrid Reward Reinforcement Learning}\label{rl}

\begin{figure}[t]
  \includegraphics[width=\columnwidth]{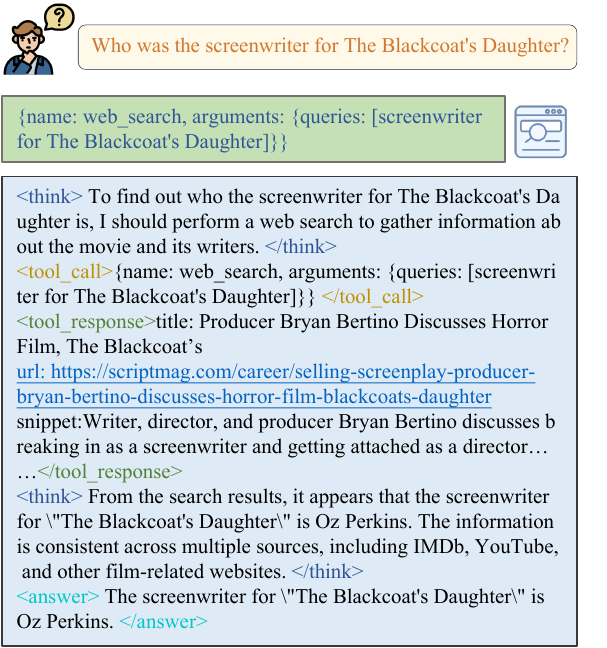}
  \caption{The illustration of a rollout that exactly matches the format.}
  \label{prompt}
\end{figure}



To encourage the model to explore diverse, high-quality rollouts, we propose hybrid reward reinforcement learning: (i) A composite reward function is designed to guide the model towards proper tool calls and the derivation of accurate answers (Section~\ref{reward}); (ii) A modified version of Group Relative Policy Optimization (GRPO) \cite{shao2024deepseekmath} is applied to improve optimization stability and bypass the need for an extra value model (Section~\ref{grpo}). The prompt for RL rollout is shown in Appendix~\ref{rl_prompt}.

\subsubsection{Reward Design}\label{reward}

We structure reward function to comprise two key components: format reward and answer reward. 

\paragraph{Format Reward.} We establish a specific template to generate rollouts, as illustrated in Figure~\ref{prompt}. According to this template, the model's \texttt{thought} for each turn is enclosed within <thinking></thinking> tags, any tool call (\texttt{action}) is placed within <tool\_call></tool\_call> tags, the corresponding tool feedback (\texttt{observation}) is enclosed by <tool\_response></tool\_response> tags, and finally, the answer is presented within <answer></answer> tags. Only when a rollout exactly matches the format, it receives a 1.0 format reward, denoted as $R_{\text{f}} = 1.0$. Otherwise, we do not further check the answer, and the rollout receives a final reward of 0.0.


\paragraph{Answer Reward.} 
When the format is strictly followed, we employ a judge model to assess the correctness of the answer (between <answer> and </answer> tags). The judge prompt is shown in Appendix~\ref{judge_prompt}. A rollout receives a 1.0 reward ($R_a = 1.0$) if the answer is correct, otherwise $R_a = 0.0$. In experiments, we also apply the F1 and recall as the answer reward for analysis, which we detail in Appendix~\ref{other_ans_reward}. 

We define the final reward as a combination of the above two rewards:
\begin{equation}
\begin{aligned}
R =
\begin{cases}  
0.5 * (R_{\text{f}}+R_{\text{a}}), & \text{if the format is correct} \\ 
0, & \text{if the format is incorrect} 
\end{cases}
\end{aligned}
\end{equation}

\subsubsection{Group Relative Policy Optimization}\label{grpo}

In this work, we adopt the Group Relative Policy Optimization (GRPO) algorithm. GRPO optimizes the current policy $\pi_\theta$ by leveraging a reference policy $\pi_{\theta_{\text{ref}}}$ along with a set of rollouts generated by an existing policy $\pi_{\theta_{\text{old}}}$. As suggested by \cite{yu2025dapo} and \cite{liu2025understanding}, we modify the original sample-level loss of GRPO into the token-level loss for better training performance. Specifically, given $G$ rollouts $\{y_i\}_{i=1}^G \sim \pi_{\theta_{\text{old}}}(\cdot | x)$(with each input $x \sim D$, where $D$ is the experience distribution), the current policy is then optimized by maximizing the following objective function:


\begin{equation}
    \begin{aligned}
    & \mathcal{J}(\theta) = \mathbb{E}_{x \sim D,  \{ y_i \}_{i=1}^G \sim \pi_{\theta_{\text{old}}}(\cdot | x) }\frac{1}{\sum_{i=1}^G |y_i|}\sum_{i=1}^G\sum_{t=1}^{|y_i|} 
    \\ & \left[ \min \left(r_{i,t} A_{i,t}, \text{clip} (r_{i,t}, 1 - \epsilon, 1 + \epsilon) A_{i,t} \right) - \beta \mathbb{D}_{\text{KL}} \right] 
    \end{aligned}
\end{equation}



where $\epsilon$ is the clipping threshold and $|y_i|$ is the length of rollout $y_i$. The $\mathbb{D}_{\text{KL}}$ represents the discrepancy of the predicted probability between the current policy $\pi_{\theta}$ and the reference policy $\pi_{ref}$. 

The advantage $A_{i,t}$ and $r_{i,t}$ are defined as follows:

\begin{equation}
\begin{aligned}
    r_{i,t}&= \frac{\pi_\theta(y_{i,t} | x, y_{i, <t})}{\pi_{\theta_{\text{old}}}(y_{i,t} | x, y_{i, <t})}\, \\
    A_{i,t}&= 
    \frac{R_i - \mathrm{mean}\bigl(\{R\}\bigr)}{\mathrm{std}\bigl(\{R\}\bigr)}
\end{aligned}
\end{equation}

where \(R_i\) represents reward for $y_i$, and
\(\mathrm{mean}(\cdot)\), \(\mathrm{std}(\cdot)\) are calculated over the batch
to normalize reward scores into advantage estimates.

\subsection{Rejection Sampling Fine-Tuning}\label{rsft}


To enhance the utilization of rollouts during the RL phase and provide a better cold-start model for the next RL iteration, we collect the rollouts in RL and use rejection sampling fine-tuning to learn high-quality and diverse samples. The following three rules are applied sequentially to ensure high-quality, diverse and multi-step rollout filtering.

\paragraph{Rule 1: High-Reward Selection (HRS).} We only retain rollouts with rewards $\geq \delta$ to ensure the high quality of training samples.



\paragraph{Rule 2: Same Query Deduplication (SQD).} 
For multiple rollouts with the same query, we retain the sample that utilizes the tools the most, to ensure the diversity of the training samples.

\paragraph{Rule 3: Multi-Calls Selection (MCS).} To enhance data utilization efficiency, we combine rollouts from the current and previous iterations. We observe that samples with multiple tool calls offer meaningful thinking and search features. As noted in Section~\ref{quality>quantity}, data quantity is prioritized over quality. Hence, we select the top k rollouts with the most tool calls for SFT in each iteration. The distribution of tool calls for RL rollouts in each iteration is presented in the Appendix~\ref{distribution_tr}.




\paragraph{Supervised Fine-Tuning.}



After obtaining the data $D_f$ filtered by the three rules, we train the base model in an SFT manner. This produces a good cold-start model for the next RL stage. Given the question $x$ and the agentic execution rollout $\mathcal{H}=(y_0, y_1,...,y_{n-1},y_n)$, where each \(y_i \in \{ \tau, \alpha, o \} \), the loss function for SFT is computed as follows:

\begin{equation}
\begin{aligned}
    L = & -\frac{1}{\sum_{i=1}^{|\mathcal{H}|} \mathbb{I}[y_i \ne o]} \\
    & \times \sum_{i=1}^{|\mathcal{H}|} \mathbb{I}[y_i \ne o] \cdot \log \pi_{\theta}(y_i \mid x, y_{<i})
\end{aligned}
\end{equation}

Here, \(  \mathbb{I}[y_i \ne o] \) filters out tokens corresponding to external feedback, ensuring that the loss is calculated only on the actions of the agent.



\section{Experiments}

\subsection{Benchmark and Evaluation Metrics}

In \Ours, we utilize the same training and testing data as DeepResearcher~\cite{zheng2025deepresearcher}. Specifically, for the training dataset, we used a distribution ratio of NQ~\cite{kwiatkowski-etal-2019-natural}:TQ~\cite{joshi-etal-2017-triviaqa}:HotpotQA~\cite{yang-etal-2018-hotpotqa}:2Wiki~\cite{ho-etal-2020-constructing} as 1:1:3:3 with a total of 80,000 samples. This includes 75\% of the samples from multi-hop scenarios, which better reflect the complex information-seeking behaviors required for deep research questions. For the evaluation dataset, we use the NQ, TQ, HotPot, and 2Wiki datasets as the in-domain evaluation set, totaling 2,048 examples. We use the Musique~\cite{trivedi2022musique}, Bamboogle~\cite{press2022measuring}, and PopQA~\cite{mallen2023llm_memorization} datasets as the out-of-domain evaluation set, totaling 1,129 examples. We utilize a judge model to evaluate the correctness of the model's response.


\subsection{Implementation Details}

We utilize Qwen2.5-7B-Instruct\footnote{https://huggingface.co/Qwen/Qwen2.5-7B-Instruct}~\cite{qwen2.5} as our backbone.
During the RL training phase, each sample undergoes 16 rollouts with a training batch size of 128, a learning rate of 1e-6, a maximum search count of 10, and a temperature of 1.0. 
The training epoch is set to 1.
We utilize Qwen2.5-72B-Instruct\footnote{https://huggingface.co/Qwen/Qwen2.5-72B-Instruct} as our judge model. 
We split the training data into $N=10$ parts, and 8,000 samples are consumed for RL training in each iteration. In the data filter process, we set k to 2000 to select samples with the highest number of tool calls, $\delta$ to 0.7 to select samples with a reward exceeding 0.7. 
In the RSFT training phase, we employ  Zero-3 offload\cite{aminabadi2022deepspeedinferenceenablingefficient}, with a batch size of 64, a learning rate of 3e-6, and the training epoch set to 1. 

\subsection{Baselines}

To evaluate the effectiveness of our approach, we compare it with the following baseline methods:

\begin{itemize}
    \item[•] CoT: This baseline generates answers using Chain-of-Thought reasoning without depending on any external reference context.
    \item[•] RAG: This method integrates CoT reasoning with retrieved reference context to assist in the generation.
    \item[•] Search-o1~\cite{Search-o1} + Web Search: A multi-step reasoning baseline where the model is permitted to generate search queries and send real-time search requests via APIs, accessing URLs to browse web pages. The model can then generate answers based on the content of these web pages.
    \item[•] Search-r1~\cite{jin2025search}: An RL-based fine-tuning strategy. During both the training and inference stages, it utilizes a retriever to access information from Wikipedia. We consider two setups: using Qwen2.5-7B-base\footnote{https://huggingface.co/Qwen/Qwen2.5-7B} or Qwen2.5-7B-Instruct as the initial actor models, respectively.
    \item[•] R1-Searcher~\cite{R1-searcher}: Unlike Search-r1, when given a search query, it searches Bing and answers questions by summarizing the first three pages of the search results.
    \item[•] DeepResearcher~\cite{zheng2025deepresearcher}: Unlike R1-Searcher, DeepResearcher does not restrict its search to a specific domain and allows for autonomous selection of URLs rather than mandatorily summarizing the top three search results.
    \item[•] DeepResearcher + Model-Based Reward (MBR): The standard DeepResearcher uses the F1 score as a reward. For fair comparison with our method, we also use Qwen2.5-72B-Instruct as the judge model for the response reward. For simplicity, we denote this baseline as  DeepResearcher$^{*}$.
    \item[•] RLSearch: An RL-only baseline of our framework. It is only trained via RL with the same raw data and the same hyperparameters.
\end{itemize}
All the baselines use Qwen2.5-7B-Instruct as the backbone unless specifically mentioned.



\subsection{Main Results}


\begin{table*}
  \centering
  \resizebox{\textwidth}{!}
  {\begin{tabular}{lc|lllll|lllll}
    \toprule
    \multirow{2}{*}{\textbf{Method}}    
    & \multicolumn{1}{c|}{\textbf{Inference}} & \multicolumn{5}{c|}{\textbf{In Domain}} & \multicolumn{4}{c}{\textbf{Out of Domain}} \\
    & \multicolumn{1}{c|}{\textbf{Environment}} & \textbf{NQ} & \textbf{TQ} & \textbf{Hotpot} & \multicolumn{1}{c}{\textbf{2Wiki}} & \textbf{Avg} & \textbf{Musique} & \textbf{Bamb} & \textbf{PopQA} & \textbf{Avg} \\
    \midrule

    \multicolumn{11}{c}{\cellcolor{babycyan} \textbf{\textit{Prompt Based}}} \\    

    CoT$^\dagger$ & Local RAG & 32.0 & 48.2 & 27.9 & 27.3 & 33.9 & 7.4 & 21.6 & 15.0 & 14.7 \\
    CoT+RAG$^\dagger$ & Local RAG & 59.6 & 75.8 & 43.8  & 24.8 & 51.0 & 10.0 & 27.2 & 48.8 & 28.7 \\
    Search-o1$^\dagger$ & Web Search & 55.1 & 69.5 & 42.4 & 37.7 & 51.2 & 19.7 & 53.6 & 43.4 & 38.9 \\
    \midrule

    \multicolumn{11}{c}{\cellcolor{aliceblue} \textbf{\textit{Training Based}}} \\
    

    Search-r1-base$^\dagger$ & Local RAG & 60.0 & 76.2 & 63.0 & 47.9 & 61.8 & 27.5 & 57.6 & 47.0 & 44.0 \\
    Search-r1-instruct$^\dagger$ & Local RAG & 49.6 & 49.2 & 52.5 & 48.8 & 50.0 & 28.3 & 47.2 & 44.5 & 49.5 \\
    R1-Searcher$^\dagger$ & Web Search & 52.3 & 79.1 & 53.1 & 65.8 & 62.6 & 25.6 & 65.6 & 43.4 & 44.9 \\
    DeepResearcher$^\dagger$ & Web Search & 61.9 & 85.0 & 64.3 & 66.6 & 69.5 & 29.3 & 72.8 & 52.7 & 51.6  \\

    $\text{DeepResearcher}^{*}$ & Web Search & 66.4 & 86.0 & 65.4  &  75.0 & 73.2 & 29.0 & 71.7 & 50.2 & 50.3 \\

    
    RLSearch-ite1 & Web Search & 68.5 & 86.3 & 66.7 & 76.4 & 74.5 & 30.4 & 74.2 & 50.4 & 51.6 \\
    RLSearch-ite2 & Web Search & 69.3 & 87.7 & 66.8 & 75.5 & 74.9 & 33.5 & 73.6 & 51.1 & 52.7 \\
    RLSearch-ite3 & Web Search & 69.8 & 88.4 & 65.0 & 71.8 & 73.8 & 30.8 & 77.0 & 51.8 & 53.2 \\
    \midrule
    \multicolumn{11}{c}{\cellcolor{lightgray} \textbf{\textit{Ours}}} \\
    \Ours -ite1 & Web Search & 68.5 & 87.4 & 65.4 & 75.6 & 74.2 & 29.3  & 74.0 & 51.2  & 51.5
    \\
    \Ours -ite2 & Web Search & 69.4 & 86.3 & 66.3 & \textbf{78.5} & 75.1 & 31.6 & 76.5 & \textbf{52.8} & 53.6 \\
    \Ours -ite3 & Web Search & \textbf{71.0} & \textbf{89.5} & \textbf{67.7} & 76.4 & \textbf{76.2} & \textbf{33.8} & \textbf{77.1} & 50.3 & \textbf{53.7} \\
    \bottomrule
  \end{tabular}}
  \caption{Main results on seven multi-hop question answering (MHQA) benchmarks. All the results labelled with $^\dagger$ are taken from \citet{zheng2025deepresearcher}.
  }
  \label{MainResults}
\end{table*}

The results of EvolveSearch for 3 iterations and other baselines are presented in Table~\ref{MainResults}.

\paragraph{\Ours\ consistently outperforms all baselines within training domains.} \Ours\ achieves the highest performance across all datasets within the four domains, significantly surpassing all baselines on the NQ and 2Wiki datasets. While DeepResearcher + MBR demonstrates comparable performance on the NQ and HotpotQA datasets, it is noteworthy that DeepResearcher + MBR utilizes model-based reward, which results in performance significantly higher than when using F1 as a reward during training. Therefore, using F1 as a reward tends to shorten the model's response, thereby affecting the overall quality of its responses.

\noindent \textbf{\Ours\ demonstrates impressive generalization capabilities in out-of-domain scenarios.} It consistently surpasses all baseline methods across three out-of-domain datasets. This indicates that \Ours\ allows the model to effectively acquire reasoning skills that can be applied broadly, instead of just adjusting to specific training data.

\noindent \textbf{A good cold-start model is crucial.} We select high-quality and diverse reasoning rollouts during the RL phase to employ Rejection Sampling Fine-Tuning (RSFT) for obtaining a better initial policy model.
From the results, especially the comparison between the three iterations of \Ours\ and the RLSearch baseline, we can conclude that a good cold-start model can further enhance the model's potential and stability in RL training, thereby consistently improving the performance.

\section{Analysis}



\begin{figure}[t]
  \includegraphics[width=\columnwidth]{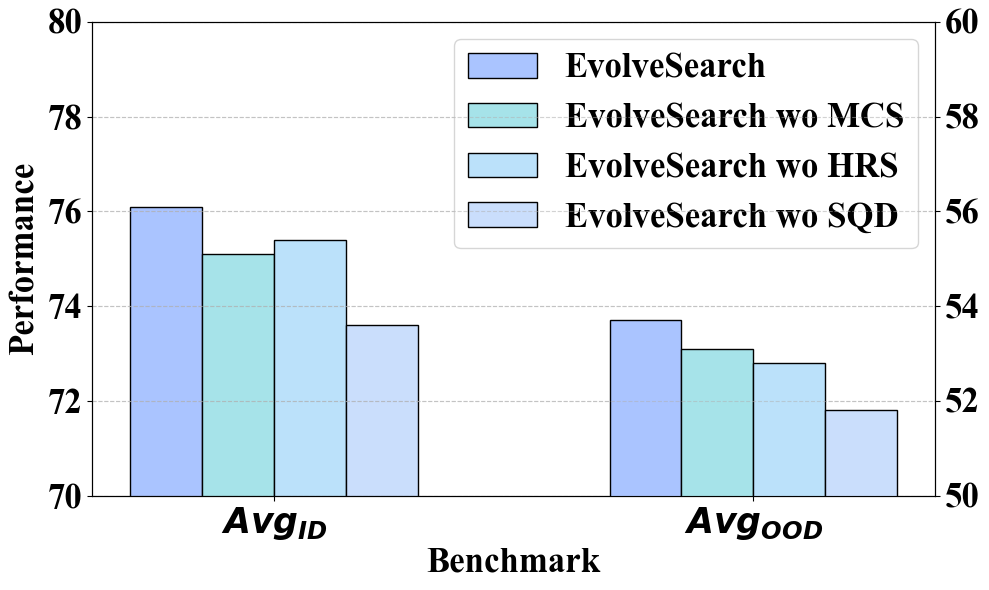}
  \caption{The impact of different data filtering rules on performance.}
  \label{DFP}
\end{figure}


\noindent \textbf{Data filtering plays a vital role.}
To demonstrate the significance of data filtering rules, we report the model's average in-domain and out-of-domain performance across different filtering rules.
For fair comparison, we randomly select only 2000 SFT samples for each experiment, followed by using the additional 8000 samples for RL training. 
As shown in Figure~\ref{DFP}, each data filtering rule is essential. 
Specifically, filtering out multi-call data aims to enhance the model's initial multi-step reasoning capabilities, filtering data with high rewards ensures data accuracy, and filtering data with different queries increases diversity, thereby comprehensively improving the model's performance.

\begin{table}
  \centering
  \resizebox{\linewidth}{!}
  {\begin{tabular}{cccc}
    \toprule
    \textbf{Training Reward} & \textbf{Method} & $\text{\textbf{AVG}}_\text{ID}$ & $\text{\textbf{AVG}}_\text{OOD}$ \\
    \midrule
    \multirow{4}{*}{Recall} & DeepResearcher & 65.5 & 52.8 \\
    & \Ours -ite1 & 68.8 & 55.8 \\
    & \Ours -ite2 & 69.6 & 56.6 \\
    & \Ours -ite3 & \textbf{69.9} & \textbf{58.8} \\
    \midrule
    \multirow{4}{*}{F1}  & DeepResearcher & 61.2 & 50.8 \\
    & \Ours -ite1 & 61.6 & 51.4 \\
    & \Ours -ite2 & \textbf{62.2} & 51.0  \\
    & \Ours -ite3 & 62.1 & \textbf{52.1} \\
    \midrule
    \multirow{4}{*}{Judge Model}  & DeepResearcher & 73.2 & 50.3 \\
    & \Ours -ite1 & 74.2 & 51.5 \\
    & \Ours -ite2 & 75.1 & 53.6 \\
    & \Ours -ite3 & \textbf{76.2} & \textbf{53.7} \\
    \bottomrule
  \end{tabular}}
  \caption{Performance comparison of different training rewards during the training phase.}
  \label{DRM}
\end{table}

\paragraph{\Ours\ remains effective with different training rewards.} To evaluate the effectiveness of our method when employing different answer rewards, we replace the answer reward component with three common and widely recognized metrics: Recall, F1 Score, and Model-Based Reward. We utilize these metrics to compare the performance of our method against the baseline. To ensure consistency between training and testing, the same answer evaluation metric is used for both. The experimental results, presented in Table~\ref{DRM}, demonstrate that our method consistently outperforms the baseline when using each of these different answer reward metrics on seven benchmarks. Furthermore, we observe that as the number of iterations of our method increases, the model's performance on both in-domain (ID) and out-of-domain (OOD) datasets gradually improves.


\begin{table}
  \centering
  \resizebox{\linewidth}{!}
  {\begin{tabular}{lccc}
    \toprule
    \textbf{Method} & \textbf{Judge Model} & $\text{\textbf{AVG}}_\text{ID}$ & $\text{\textbf{AVG}}_\text{OOD}$ \\
    \midrule
    \multirow{3}{*}{$\text{DeepResearcher}^{*}$}  & DeepSeek-V3 & 66.1 & 38.0 \\
    & chatgpt-4o-latest & 71.0 & 42.5 \\
    & grok-3 & 74.5 & 45.9 \\
    \midrule
     \multirow{3}{*}{$\text{\Ours-ite1}$}  & DeepSeek-V3 & 66.3 & 39.8 \\
    & chatgpt-4o-latest & 71.8 & 43.9 \\
    & grok-3 & 74.7 & 48.0 \\
    \midrule
    \multirow{3}{*}{$\text{\Ours-ite2}$}  & DeepSeek-V3 & 67.4 & 41.1 \\
    & chatgpt-4o-latest & 72.6 & 45.0 \\
    & grok-3 & 75.3 & 48.7 \\
    \midrule
    \multirow{3}{*}{$\text{\Ours-ite3}$}  & DeepSeek-V3 & 67.4 & 41.6 \\
    & chatgpt-4o-latest & 72.7 & 45.3 \\
    & grok-3 & 75.8 & 49.3 \\
    \bottomrule
  \end{tabular}}
  \caption{Comparison of model performance using different judge models.}
  \label{DJM}
\end{table}

\paragraph{EvolveSearch still demonstrates superior performance across different judge models.} To further verify our approach's effectiveness, we utilize different judge models to evaluate the model's response. In Table~\ref{DJM}, we select three well-known LLMs, DeepSeek-V3~\cite{deepseekai2024deepseekv3technicalreport}, chatgpt-4o-latest\footnote{https://openai.com}, and grok-3\footnote{https://x.ai} as the judge model. Although their performance is slightly lower than the trained judge model, the improvement trend relative to the baseline remains consistent. After the first iteration of training, the model outperforms the baseline in both in-domain and out-of-domain benchmarks. As the iteration increases, the model's performance gradually improves, further demonstrating the effectiveness of the method.

\begin{figure}[t]
  \includegraphics[width=\columnwidth]{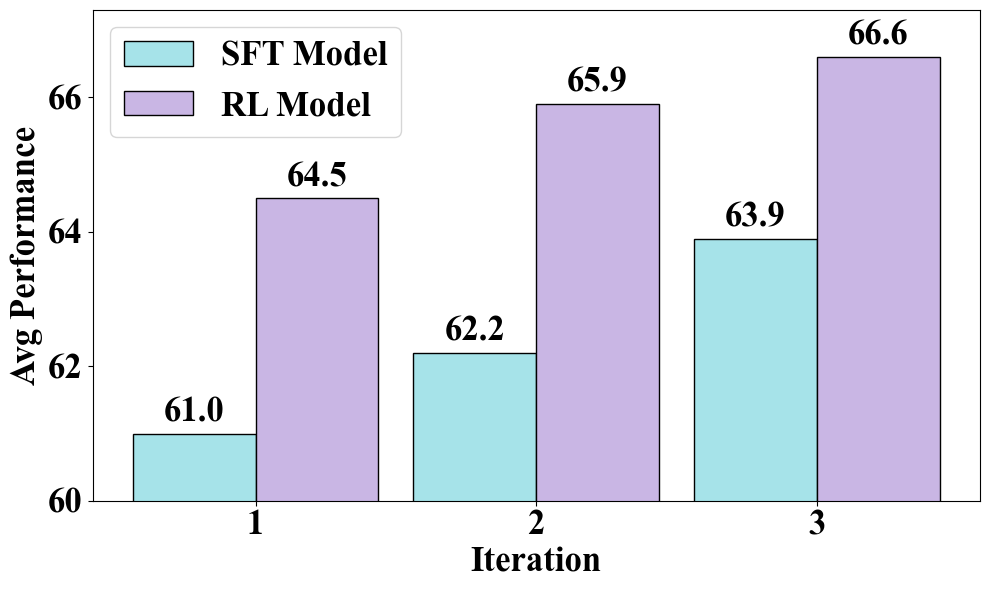}
  \caption{Performance of SFT model and RL model at different iterations.}
  \label{Improvement}
\end{figure}


\paragraph{The performance of the SFT Model and RL Model improves as the number of iterations increases.} Figure~\ref{Improvement} presents the performance of the SFT model and RL model across different iterations. We observe that as iterations grow, not only does the RL Model exhibit significant improvements across seven different benchmarks, but the SFT Model also shows considerable enhancement. This confirms the high quality of our chosen data and demonstrates the effectiveness of EvolveSearch.

\begin{figure}[t]
  \includegraphics[width=\columnwidth]{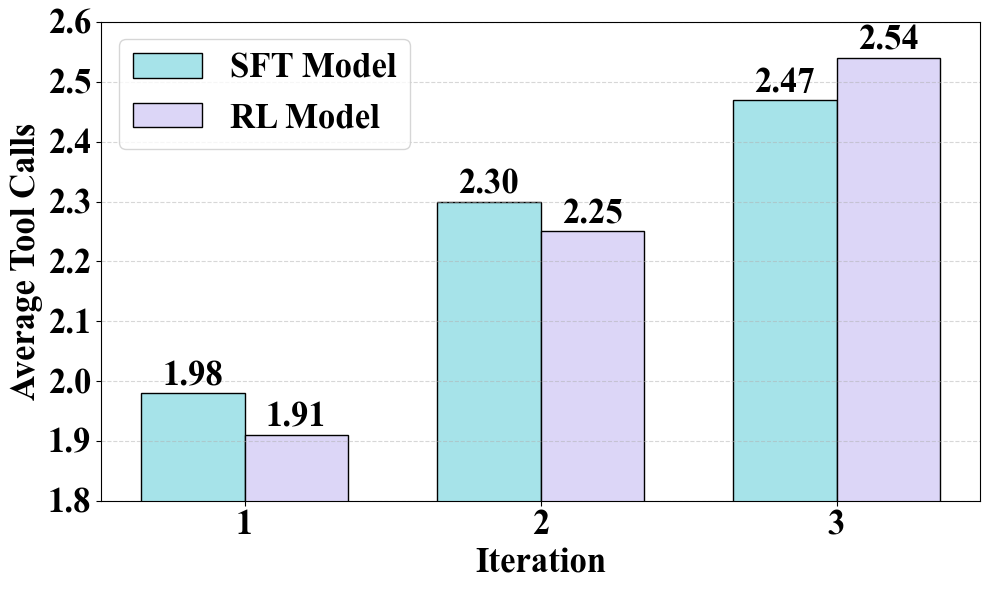}
  \caption{Average number of tool calls in the test set across different iterations.}
  \label{toolcall}
\end{figure}

\paragraph{Iterative training increases the frequency of model tool calls.} To investigate the model's tool usage on the test set throughout the entire iterative training process, we record the average number of search tool calls by the SFT model and the RL model on the test set. As shown in Figure~\ref{toolcall}, our findings reveal that, as the number of iterations increases, the model increasingly depends on the tool, leading to gathering more information. Furthermore, we note that for the majority of questions, the model requires only three calls to the search tool to reach a solution, indicating that the training data is not sufficiently challenging. A test case of the interaction between the model and the environment is presented in the Appendix~\ref{Case Study}.




\begin{table}
  \centering
  \begin{tabular}{ccc}
    \toprule
    \textbf{Training Num} & $\text{\textbf{AVG}}_\text{ID}$ & $\text{\textbf{AVG}}_\text{OOD}$ \\
    \midrule
    4000 & 75.6 & 53.3 \\
    8000 & 75.2 & 52.6 \\
    12000 & 75.4 & 53.0 \\
    16000 & 75.4 & 53.4 \\
    \bottomrule
  \end{tabular}
  \caption{The impact of different data volumes on model performance during the RSFT phase.}
  \label{DQDQ}
\end{table}

\paragraph{Data quality is more important than data quantity.}\label{quality>quantity} 

To investigate the impact of SFT training data volume on model performance, we conduct controlled experiments using progressively scaled datasets. Based on historical RL training rollouts, we remove the filtering on the number of web search tool calls. SFT is performed at four different scales: 4,000, 8,000, 12,000, and 16,000 samples. All experiments subsequently undergo an identical RL training phase using a fixed 8,000 question-answering samples. The results are presented in Table~\ref{DQDQ}, which show that merely increasing data volume does not necessarily enhance performance, which indicates that the quality of data is more important than the quantity.

\section{Related Work}


\paragraph{Search Agent.} Current methods often rely on manually designed workflows to guide large language models (LLMs) in interacting with external knowledge sources~\cite{wu2025webwalkerbenchmarkingllmsweb}. Recent studies like OpenResearcher~\cite{zheng2024openresearcher}, IterDRAG~\cite{yue2025inference}, AirRAG~\cite{feng2025airrag}, and others have improved search capabilities using these detailed workflows. However, these approaches are limited by their dependence on human-crafted prompts and interaction patterns. Recent developments about SFT for Retrieval-Augmented Generation (RAG) have become a preferred method over manual optimization~\cite{yu2024auto}. For example, CoRAG~\cite{wang2024corag} utilizes Monte Carlo Tree Search (MCTS) to select optimal document blocks under budget constraints but faces high computational costs and limited generalization due to reliance on supervised signals. Reinforcement Learning (RL) presents an end-to-end approach to enhance large language models' capabilities, improving reasoning skills significantly by late 2024~\cite{ouyang2022training, shao2024deepseekmath}. Recent research explores RL for external knowledge retrieval, with systems like Search-R1~\cite{jin2025search}, ReSearch~\cite{chen2025research}, and R1-Searcher~\cite{R1-searcher} evolving beyond predefined cues to models that autonomously develop reasoning during retrieval. However, these methods often converge quickly, resulting in low data efficiency and limited performance gains.

\paragraph{Self-Evolution.} Large language models (LLMs) have shown the capability to annotate datasets without relying on human-annotated labels, enabling low-resource training for other LLMs. In typical setups, a larger model, the teacher, generates labels for a smaller model, the student, in a process known as \textit{context distillation}. Various algorithms can be employed, such as conventional supervised fine-tuning (SFT)~\cite{stanford_alpaca,hsieh2023distilling}, in-context learning~\cite{krishna2024post}, and preference optimization~\cite{tunstall2023zephyr,meta_llama_3}. Self-evolution methods remove the necessity for a larger LLM, reducing computational demand and API costs. Recent research has shown this approach is viable using an unlabeled dataset with a few examples for; context~\cite{huang2022large,tian2023fine}. For instance, ~\cite{he2019revisiting} uses a small labeled dataset for initial fine-tuning before applying the trained generator to annotate the unlabeled data, similar strategies are employed for rationalization tasks in ~\cite{jie2024plausible}. ~\cite{meng2022generating} enhances labeled datasets with additional samples, though this is limited to classification. Our approach focuses on the open web search domain, combining SFT and RL to enhance search capabilities without requiring any external human-annotated reasoning data.

\section{Conclusion}

In an era where current search agents have surpassed the capabilities of most humans, we propose \Ours, a novel iterative self-evolution framework that synergistically combines RL with SFT to enhance web search capabilities without any external human-annotated reasoning rollouts. 
Extensive experiments on multiple MHQA benchmarks demonstrate that \Ours\ consistently improves performance with each iteration, ultimately achieving an average accuracy improvement of 4.7\% over SOTA methods on seven benchmarks, paving the way to self-evolution and self-improvement in open web search domains.

\section*{Limitations}
EvolveSearch relies on iterative collection and filtering of the rollouts during RL training for SFT, which adds to the computation cost of the whole training process. Designing a streaming rollout filtering system with higher sample efficiency is one possible way to minimize the impact of extra computation.

In this work, the tool call is limited to web search, so the performance of other tools remains unknown. The investigation of multiple tools is left for future research.

\bibliography{custom}

\clearpage
\appendix

\section{RL Prompt}
\label{rl_prompt}
An example of rollout in the RL phase is shown below.

\begin{tcolorbox}[
enhanced jigsaw,
breakable,
pad at break*=1mm,
colback=white!95!gray,
colframe=gray!50!black,
title={Prompts for RL Rollout}]
\small
\begin{lstlisting}[breaklines=true, xleftmargin=0pt, breakindent=0pt, columns=fullflexible]
A conversation between User and Assistant. The user asks a question, and the assistant solves it by calling one or more of the following tools.
<tools>
{
  "name": "web_search",
  "description": "Utilize the web search engine to retrieve relevant information based on multiple queries.",
  "parameters": {
    "type": "object",
    "properties": {
      "queries": {
        "type": "array",
        "items": {
          "type": "string",
          "description": "The search query."
        },
        "description": "The list of search queries."
      }
    },
    "required": ["queries"]
  }
}
</tools>

The assistant starts with one or more cycles of (thinking about which tool to use -> performing tool call -> waiting for tool response), and ends with (thinking about the answer -> answer of the question). The thinking processes, tool calls, tool responses, and answer are enclosed within their tags. There could be multiple thinking processes, tool calls, tool call parameters and tool response parameters.

Example response:
<think> thinking process here </think>
<tool_call>
{"name": "tool name here", "arguments": {"parameter name here": parameter value here, "another parameter name here": another parameter value here, ...}}
</tool_call>
<tool_response>
{"name": "tool name here", "content": {"result name here": result value here, "another result name here": another result value here, ...}}
</tool_response>
<think> thinking process here </think>
<tool_call>
{"name": "another tool name here", "arguments": {...}}
</tool_call>
<tool_response>
{"name": "another tool name here", "content": {...}}</tool_response>
(more thinking processes, tool calls and tool responses here)
<think> thinking process here </think>
<answer> answer here </answer>

User: (*@ \textbf{\{INPUT QUERY\}} @*)"
\end{lstlisting}
\end{tcolorbox}
\section{Judgement Prompt}\label{judge_prompt}
The prompt for answer judgement in our work is based on \citet{wei2024measuring}.
The detailed prompt is shown below.

\begin{tcolorbox}[
enhanced jigsaw,
breakable,
pad at break*=1mm,
colback=white!95!gray,
colframe=gray!50!black,
title={Prompts for Answer Judgement}]
\small
\begin{lstlisting}[breaklines=true, xleftmargin=0pt, breakindent=0pt, columns=fullflexible]
Please evaluate whether the model's response is correct based on the given question, standard answer, and the model's predicted answer. Your task is to rate the result as: (*@\textbf{Correct} @*) or (*@\textbf{Incorrect} @*).

(*@\textbf{Correct Response} @*)

Here are examples of Correct responses:

Question: What are Barack Obama's children's names?
Standard Answer: Malia Obama and Sasha Obama

Model Prediction 1: Malia Obama and Sasha Obama
Model Prediction 2: Malia and Sasha
Model Prediction 3: Most people would say Malia and Sasha, but I'm not sure and need to confirm.
Model Prediction 4: Barack Obama has two daughters, Malia Ann and Natasha Marian, but they are commonly known as Malia Obama and Sasha Obama.
Model Prediction 5: Barack Obama's children

These responses are Correct because:
They fully include the important information from the standard answer.
They do not contain any information that contradicts the standard answer.
Only the semantic content is considered; language (English or Chinese), case, punctuation, grammar, and order are not important.
The presence of vague statements or guesses is acceptable, as long as the standard answer is included and there is no incorrect or contradictory information.

(*@\textbf{Incorrect Response} @*)

Here are examples of Incorrect responses:

Question: What are Barack Obama's children's names?
Standard Answer: Malia Obama and Sasha Obama

Model Prediction 1: Malia
Model Prediction 2: Malia, Sasha, Susan, and Sasha Obama or Malia Obama, or Natasha Marian, or Einstein
Model Prediction 3: Although I don't know their exact names, I can say that Barack Obama has two children.
Model Prediction 4: You might be thinking of Bessie and Olivia. But you should check the latest references for detailed information. Is that the correct answer?
Model Prediction 5: Barack Obama's children

These responses are Incorrect because:
They contain factual statements that contradict the standard answer.
The answer is empty, restates the question.
The answer lists multiple answers, restates the answer.

(*@\textbf{Special Notes} @*)

Please note the following: 
The standard answer may contain multiple aspects of the question's response, and within the same aspect, there may be multiple different descriptions, all of which are correct and are given within the same parentheses, connected by commas. For example, consider the question ''What is the name of the social media platforms purchased by Elon Musk?'':
Predicted answers ''Twitter,'' ''Twitter, X,'' and ''X'' are all Correct.
For standard answers that contain responses to multiple aspects of the question, the model must provide answers to all aspects to be considered correct; otherwise, it is directly judged as Incorrect. There is no such output as (*@ \textbf{Partially Correct}. @*) These answers will be given in different parentheses. For example, consider the question ''Who are the original members of the band The Beatles?'':
Predicted answers ''John Lennon, Paul McCartney, George Harrison, Ringo Starr'' that include all answers are considered Correct.
Predicted answers like ''John Lennon, Paul McCartney'' that do not include all answers are considered Incorrect.

(*@\textbf{Additional Guidelines} @*)
Also, pay special attention to the following: 
For questions with numerical standard answers, the predicted answer should match the standard answer. For example, consider the question ''What is the total length of the Jinshan Railway Huangpujiang Special Bridge in meters?'':
Predicted answers ''3518,'' ''3518.1,'' and ''3518.17'' are all Correct.
Predicted answers ''3520'' and ''3600'' are all Incorrect.
If the model's prediction does not directly answer the question and attempts to bypass or fails to directly provide the standard answer, it is considered an Incorrect answer.
If the standard answer contains more information than the question, the predicted answer only needs to include the information mentioned in the question.
If it is obvious from the question that the predicted answer has omitted information, it is considered Correct.
If it is clear that different translation versions of a name refer to the same person, they are also considered Correct.
You should focus more on the match between the standard answer and the model's prediction, rather than whether the standard answer is correct.

(*@\textbf{Example Question} @*)
Here is a new example question. Please rate the predicted answer as one of the following: 
Question: {question} 
Standard Answer: {target} 
Predicted Answer: {predicted answer} 
Only return the option represented by Correct or Incorrect, that is, only return A or B, without adding any other text.

\end{lstlisting}
\end{tcolorbox}




\section{Distribution of RL Rollouts}
\label{distribution_tr}
\begin{figure}[h]
  \includegraphics[width=\columnwidth]{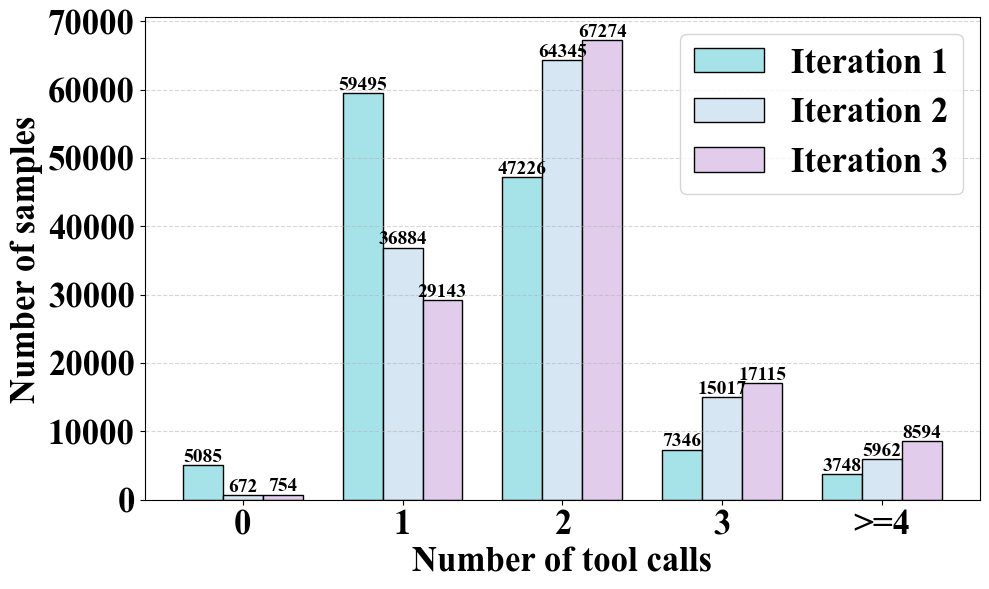}
  \caption{The distribution of tool calls for RL rollouts. As the iteration increases, the average number of tool calls also increases, and the number of trivial rollouts ($\leq 1$ tool call) significantly decreases. This indicates that the rollouts gain in both diversity and quality during training.}
  \label{DataNum}
\end{figure}

The distribution of RL Rollouts is shown in Figure~\ref{DataNum}.

\begin{algorithm}
    \caption{The iterative workflow of \textbf{EvolveSearch}.}
    \label{Ours_algorithm}
    
    \textbf{Input}: The number of iterations $N$, raw data $\mathcal{RD}=\{\{d_{1},a_{1}\},...,\{d_{n},a_{n}\}\}$, a base model $\mathcal{M}$. \\
    \textbf{Output}: RL model $\mathbf{M}_{rN}$ in the last iteration.
    \begin{algorithmic}[1] 

        \State $\mathcal{RD}^\star \leftarrow \{\mathcal{RD}^\star_1,..., \mathcal{RD}^\star_N\}$ \textcolor{cyan}{\Comment{Divide the raw data evenly into $N$ parts}} 

        \State $\mathcal{DP} \leftarrow \{\}$  \textcolor{cyan}{\Comment{Initialize data pool}}

        \For{$i \gets 1 ... N$}

            \State $\mathcal{D} \leftarrow GET(\mathcal{DP})$ \textcolor{orange}{\Comment{Get all data}}

            \State $\mathcal{FD} $ $\leftarrow \mathcal{F}(D) $ \textcolor{orange}{\Comment{Filter data by three rules}}

            \State $\mathcal{M}_{s} \leftarrow SFT (\mathcal{M},\mathcal{FD})$\textcolor{blue}{\Comment{Get SFT model}}
    
            \State $\mathcal{M}_{ri} \leftarrow RL (\mathcal{M}_{s},\mathcal{RD}^\star_i)$ \textcolor{blue}{\Comment{Get RL model}}
    
            \State $ \mathcal{D}^r_i \leftarrow \{\{d_{i1}, t_{i1},a_{i1}\},...,\{d_{ik},t_{ik},a_{ik}\}\}$ \textcolor{orange}{\Comment{Obtain $i$-th RL rollout data}}
    
            \State $\mathcal{DP} \leftarrow \mathcal{DP}\  \cup\ \mathcal{D}^r_i $ \textcolor{orange}{\Comment{Combine $\mathcal{D}$ with $\mathcal{D}^r_i$}}
            
        \EndFor

         \State \Return $\mathcal{M}_{rN}$

    \end{algorithmic}
\end{algorithm} 

\section{The Iterative Algorithm Workflow}
\label{Algorithm}

The overall workflow framework of \textbf{EvolveSearch} is shown in Algorithm~\ref{Ours_algorithm}.

\section{Case Study}
\label{Case Study}

A test case of \textbf{EvolveSearch} is presented below. In this case, the model demonstrates abilities such as critical thinking and refinement through multiple tool calls, proving the effectiveness of our training method.

\begin{tcolorbox}[
enhanced jigsaw,
breakable,
pad at break*=1mm,
colback=white!95!gray,
colframe=gray!50!black,
title={EvolveSearch Case}]
\small
\begin{lstlisting}[breaklines=true, xleftmargin=0pt, breakindent=0pt, columns=fullflexible]
Query: What was the name changed to after Allison J71 was modified to a thrust of 7400 lbf?
(*@ \textcolor{blue}{<think>} @*)
To find the name change after Allison J71 was modified to a thrust of 7400 lbf, I need to first determine which model was created with this modification.
(*@ \textcolor{blue}{</think>} @*)
(*@ \textcolor{orange}{<tool\_call>} @*)
{\"name\": \"web_search\", \"arguments\": {\"queries\": [\"Allison J71 modified to 7400 lbf thrust model name\"]}}
(*@ \textcolor{orange}{</tool\_call>} @*)
(*@ \textcolor{cyan}{<tool\_response>} @*)
...
(*@ \textcolor{cyan}{</tool\_response>} @*)
(*@ \textcolor{blue}{<think>} @*)
The web search detailed various F-89 Scorpion versions but didn't mention the Allison J71 modified to 7400 lbf thrust. I'll refine the search to find the specific model name linked to this engine modification.
(*@ \textcolor{blue}{</think>} @*)
(*@ \textcolor{orange}{<tool\_call>} @*)
{\"name\": \"web_search\", \"arguments\": {\"queries\": [\"Allison J71 modified to 7400 lbf thrust model name F-89 Scorpion\"]}}
(*@ \textcolor{orange}{</tool\_call>} @*)
(*@ \textcolor{cyan}{<tool\_response>} @*)
...
(*@ \textcolor{cyan}{</tool\_response>} @*)
(*@ \textcolor{blue}{<think>} @*)
The web search results don't specify the model name change for the Allison J71 modified to 7400 lbf thrust but indicate it's developed from the J35 engine, particularly the J35-A-23 variant. Since the F-89 Scorpion used the J71, it's likely that the thrust modification was part of an engine upgrade rather than a model name change. I will try to identify if any specific F-89 Scorpion model included this engine modification.
(*@ \textcolor{blue}{</think>} @*)
(*@ \textcolor{orange}{<tool\_call>} @*)
{\"name\": \"web_search\", \"arguments\": {\"queries\": [\"F-89 Scorpion model with Allison J71 7400 lbf thrust\"]}}
(*@ \textcolor{orange}{</tool\_call>} @*)
(*@ \textcolor{cyan}{<tool\_response>} @*)
...
(*@ \textcolor{cyan}{</tool\_response>} @*)
(*@ \textcolor{blue}{<think>} @*)
The web search suggests that the Allison J35 engine with 7,400 lbf wet thrust was used in the F-89 Scorpion. Specifically, the F-89C-25-NO block adopted the J35-A-33 engine, providing a dry thrust of 5,600 lbf and 7,400 lbf with afterburner. This implies the sought name change might be tied to this engine upgrade in the F-89C series, possibly altering the model designation to F-89C-25-NO.
(*@ \textcolor{blue}{</think>} @*)
(*@ \textcolor{purple}{<answer>} @*)
The name change after modifying the Allison J71 to a thrust of 7400 lbf likely relates to the F-89C-25-NO production of the F-89 Scorpion, which used the Allison J35-A-33 engine with the needed thrust specifications.
(*@ \textcolor{purple}{</answer>} @*)

\end{lstlisting}
\end{tcolorbox}

    









\section{Other Answer Reward}\label{other_ans_reward}
For F1, we directly define $R_{\text{a}}$ as the exact F1 score of the predicted answer and the gold answer, i.e., $R_{\text{a}} = F1(\text{pred}, \text{gold}) \in [0,1]$.

For recall, through some preliminary experiments, we found that a hard label is more helpful in RL rather than a soft score. Therefore, we define $R_{\text{a}}$ as follows:
\begin{equation*}
\begin{aligned}
R_{\text{a}} =
\begin{cases}  
1.0, & \text{if the recall score = 1.0} \\ 
0.0, & \text{if the recall score < 1.0} 
\end{cases}
\end{aligned}
\end{equation*}
\end{document}